\documentclass{article}

\usepackage{microtype}
\usepackage{graphicx}
\usepackage{subfigure}
\usepackage{booktabs} 

\usepackage{algorithmic}
\usepackage{algorithm}
\usepackage{natbib}
\usepackage{longtable} 
\usepackage{bchart}
\usepackage{tabularx}
\usepackage{xcolor}
\usepackage{multirow}
\usepackage{caption}
\usetikzlibrary{patterns}

\definecolor{forestgreen}{rgb}{0.4, 0.69, 0.2}
\definecolor{harvardcrimson}{rgb}{0.79, 0.0, 0.09}
\definecolor{hanblue}{rgb}{0.27, 0.42, 0.81}

\usepackage{hyperref}

\usepackage[accepted]{icml2020}

\icmltitlerunning{Sparse Communication for Training Deep Networks}

\begin{document}

\twocolumn[
\icmltitle{Sparse Communication for Training Deep Networks}



\icmlsetsymbol{equal}{*}

\begin{icmlauthorlist}
\icmlauthor{Negar Foroutan Eghlidi}{to}
\icmlauthor{Martin Jaggi}{to}

\end{icmlauthorlist}

\icmlaffiliation{to}{EPFL, Lausanne, Switzerland}

\icmlcorrespondingauthor{Martin Jaggi}{martin.jaggi@epfl.ch}
\icmlcorrespondingauthor{Negar Foroutan Eghlidi}{negar.foroutan@epfl.ch}

\icmlkeywords{Machine Learning, ICML}

\vskip 0.3in
]



\printAffiliationsAndNotice{}  

\begin{abstract}

Synchronous stochastic gradient descent~(SGD) is the most common method used for distributed training of deep learning models.
In this algorithm, each worker shares its local gradients with others and updates the parameters using the average gradients of all workers.
Although distributed training reduces the computation time, the communication overhead associated with the gradient exchange forms a scalability bottleneck for the algorithm.
There are many compression techniques proposed to reduce the number of gradients that needs to be communicated.
However, compressing the gradients introduces yet another overhead to the problem.
In this work, we study several compression schemes and identify how three key parameters
affect the performance.
We also provide a set of insights on how to increase performance and introduce a simple sparsification scheme, random-block sparsification, that reduces communication while keeping the performance close to standard SGD.

\end{abstract}

\section{Introduction}

Deep neural networks~(DNNs) have recently gained large popularity due to their impressive performance on various machine learning tasks such as image and speech recognition~\cite{chiu2018state, he2016deep}, natural language processing~\cite{devlin2018bert}, and recommender systems~\cite{zhang2019deep}.
Training these networks using algorithms such as stochastic gradient descent~(SGD) is however computationally intensive despite all the improvements that have been made in hardware, training methods, and network architecture.
A large body of work has been conducted to tackle this problem through distributed training of these networks by splitting data across multiple machines and taking advantage of data parallelism~\cite{strom2015scalable, iandola2016firecaffe, goyal2017accurate, dean2012large, das2016distributed}.

Synchronous SGD is one of the most common methods used for distributed training.
In this algorithm, each worker machine calculates the gradients on its local data, then it shares the local gradients with other workers and updates the parameters using the average gradients of all workers~\cite{das2016distributed, dean2012large}.
While this algorithm significantly reduces the computation time of the training phase, the communication time needed for gradient exchange increases significantly as the number of workers increases, forming a performance bottleneck.
Moreover, with the high demand for training on edge and mobile devices for the sake of privacy and security, this problem is more prevalent than ever~\cite{mcmahan2016communication, li2020federated}.
The low network bandwidth of edge devices together with the high energy cost of wireless communication make it impractical to take advantage of distributed learning on such devices.

There are several proposals trying to address the communication bottleneck by compressing gradients (i.e., sparsification or quantization) and thus reducing the amount of data needed to be shared among the workers~\cite{lin2017deep, seide20141, stich2018sparsified, alistarh2016qsgd, renggli2018sparcml, koloskova2019decentralized}.
However, such proposals have not been widely adopted in practice because they are either too slow or do not reach the same performance due to lossy compression schemes~\cite{vogels2019powersgd}.

In this work, we study several compression schemes and identify the key parameters affecting the performance.
We also provide a set of insights on how to increase performance and introduce a simple sparsification scheme, random-block sparsification, that reduces communication while keeping the performance close to standard SGD.

\section{Background and Motivation}

There are two main approaches to compress gradients: quantization, which compresses the gradients to low-precision values~\cite{alistarh2016qsgd}, and sparsification, which reduce the communication bandwidth requirement by reducing the number of non-zero entries in the stochastic gradient~\cite{wangni2018gradient}.

Quantization-based compression schemes reduce the communication overhead by limiting the number of bits used to represent floating-point numbers.
There is a large body of work applying quantization-based schemes to minimize the communication bandwidth requirement by considering various numerical representations and various optimization objective functions~\cite{wen2017terngrad, alistarh2016qsgd, zhang2019quantized, zhou2016dorefa}.
For instance, in~\cite{seide20141}, the authors have empirically shown a single sign bit per value is enough to train DNNs if the quantization error is carried forward across mini-batches.
While these schemes are efficient and provide theoretical guarantees, in practical settings, some of the largest benefits are provided by sparsification schemes~\cite{alistarh2018convergence}.


Sparsification-based approaches reduce the communication overhead by sending only a small fraction of the entries of each gradient.
Prior work has empirically shown that convolutional and recurrent neural networks can tolerate extremely high gradient sparsity (i.e., using only 0.1$\%$ of the gradients)~\cite{lin2017deep}.
However, some heuristic approaches, such as momentum correction, local gradient clipping, momentum factor masking, and warm-up training, are needed to preserve the model's accuracy.

In \cite{strom2015scalable}, the authors proposed a sparsification method that compresses gradients by pruning the gradients that are smaller than a predefined threshold.
However, finding the optimal threshold is not easy in practice.
The authors of~\cite{stich2018sparsified} have theoretically proven that SGD converges to the optimal solution using general compression schemes~(e.g., top-k and random-k) for strongly convex and smooth problems, if error feedback is used.
While most of these efforts look at the gradient compression from theoretical point of view, there is a lack of practical analysis in this domain to examine the factors that are important in practice and affect performance.

In this work, we study several compression schemes and identify the key parameters affecting the performance.
We also provide guidelines on how to increase performance and introduce a simple sparsification scheme, random-block sparsification, that reduces communication overhead while an accuracy close to standard SGD.





\section{Methodology}

In this section we first describe our baseline sparsified SGD algorithm and later introduce the three key parameters that can affect the performance (both training execution time and test accuracy) of the algorithm.
Later in section~\ref{experiments} we analyze the effects of these parameters and present a set of insights in how to increase performance. 

Algorithm~\ref{Sparsified_SGD} gives an overview of our sparsification algorithm.
Similar to~\cite{karimireddy2019error}, error feedback is used in order to prevent convergence issues. That is, the difference between the actual and the sparsified gradient is stored locally and then added later in the next iteration.
Although only a fraction of gradient entries is sent in each iteration, eventually all are used over time.
Each worker~($w$) performs the following steps in each iteration~($t$) of the algorithm:

\begin{itemize}
  \item Calculates its local gradients~($\textbf{g}^w_t$).
  \item Adds its accumulated gradients from previous iterations~($\textbf{e}^w_t$) to the current gradients.
  \item Sparsifies the results of the previous step using a gradient compressor~($C$).
  \item Exchanges and aggregates the sparsified gradients.
  \item Updates its local accumulated gradients~($\textbf{e}^w_t$) by adding the gradients of the current iteration that were not sent to other workers.
  \item Updates the model parameters using the aggregated gradients.
\end{itemize}

\begin{algorithm}
\caption{Sparsified SGD with error-feedback}\label{Sparsified_SGD}
\begin{algorithmic}[1]
    \STATE {\textbf{Input}: $W$ workers, $\textbf{x}_0^w \in R^d$ where $ w=1,..,W$, learning rate $\gamma$, compressor~C(.)}\\
	\STATE {\textbf{Initialize}: $\textbf{e}^w_0 = 0 \in R^d$, $w=1,...,W$}
	\FOR{$t = 0,...,T-1$}
	    \FOR{$w=1,...,W$}
	    \STATE{$\textbf{g}^w_t :=$  $StochasticGradient(\textbf{x}^w_t)$}
	    \STATE{$\textbf{p}^w_t := \gamma \textbf{g}^w_t + \textbf{e}^w_t$}
	    \STATE{$\textbf{q}^w_t := C(\textbf{p}^w_t)$}
	    \STATE{$ExchangeGradients(\textbf{q}^w_t)$}
	    \STATE{$\textbf{q}_t := \sum_{w=1}^{W} \textbf{q}^w_t$}
	    \STATE{$\textbf{x}^w_{t+1} := \textbf{x}^w_t - \textbf{q}_t$}
	    \STATE{$\textbf{e}^w_{t+1} := \textbf{p}^w_t - \textbf{q}_t$}
	    \ENDFOR
	\ENDFOR
	
\end{algorithmic}
\end{algorithm}

We now introduce the three key parameters that can potentially affect the algorithm's performance.
We explore a few options for each parameter in order to understand how each parameter can affect the algorithm's performance.

The first parameter is the scope of sparsification, which indicates whether to sparsify the gradients globally or layer-wise.
In the layer-wise version, we sparsify the gradients of each network layer separately.
However, in the other case, we first concatenate the gradient vectors of all the layers and then apply the sparsification step.
We later show in Section~\ref{experiments} how this choice affects the test accuracy of the model.

The next parameter is the compressor or the spasification scheme that indicates which gradients must be pruned and which ones are sent to other workers.
We consider the following three schemes:
\begin{itemize}
    \item \textbf{Top-k Sparsification}: In this scheme, the top $k$ gradients are selected according to their values to be shared with other workers.
    Finding the top $k$ values is however a computationally expensive process.
    
    \item \textbf{Random-k Sparsification}: In this scheme, $k$ gradients are randomly selected to be shared with other workers.
    Even though this approach is simple, it suffers from the random memory accesses during both compression and decompression.
    
    \item \textbf{Block-random-k Sparsification}: This is our proposed scheme which is an optimized version of the previous scheme. In this scheme, a random gradient is first selected, and then that gradient along with its following $k-1$ values are selected to be shared with other workers.
    This approach has less overhead compared to the Random-k approach as only one random access it performed.
    
\end{itemize}

\begin{figure}%
\centering
\subfigure[Reduce]{%
\label{fig:reduce}%
\includegraphics[width=.28\columnwidth]{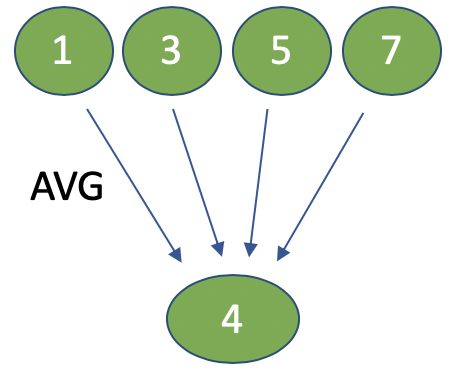}}%
\qquad
\subfigure[Gather]{

\label{fig:gather}%
\includegraphics[width=.28\columnwidth]{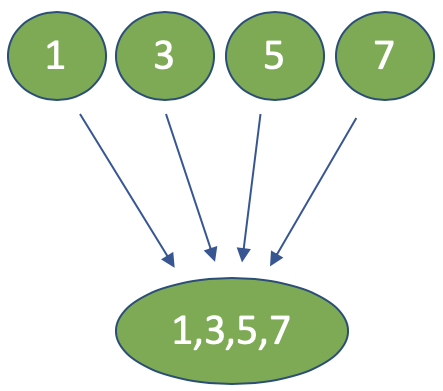}}%
\caption{Reduce and gather operations.}
\label{fig:reduce-gather}
\vspace{-15pt}
\end{figure}

Finally, the third parameter is the communication scheme used to exchange the gradients.
We consider a peer-to-peer distributed architecture in which each worker exchanges information with all the other workers in the system.
For the random-k and block-random-k schemes the workers can either use the same gradient coordinates (using the same seed state) or use different ones.
We may use different communication schemes depending on our choice to select the gradient coordinates.
\verb|allReduce| is used when all workers use the same gradient coordinates and \verb|allGather| is used when they use different coordinates.
In \verb|allReduce|, the target vectors in all workers are reduced to a single vector which is sent to all workers, while in \verb|allGather|, all the vector elements of all the workers are gathered to all the workers, meaning that all workers will have the gradient vectors of all the workers.
Figure~\ref{fig:reduce-gather} shows the reduce and gather operations from a worker's point of view, assuming the gradient vector is just one element.

\section{Evaluation}
\label{experiments}
In this section, we first describe our evaluation setup and then discuss the results.

\subsection{Experimental setup}
We implemented all our SGD algorithm using PyTorch 0.4.1~\cite{NEURIPS2019_9015} and its MPI communication backend.
The implementation is also released as part of MLBench~\cite{MLBench_benchmark}.
We train Resnet18 network on CIFAR-10~\cite{krizhevsky2009learning} dataset and employ the standard data augmentation and pre-processing schemes mentioned in~\cite{He_2016_CVPR}.
In all the experiments, we set the batch size to 128 and the algorithm runs for 300 epochs.
We set $k$ to 1\% of the gradient vector size for the three sparsification schemes.

We set the initial learning rate to $\gamma=0.1$ for our standard SGD and SGD with layer-wise compression implementations,
and $\gamma = 0.01$ for SGD with global compression implementation.
The learning rate is then divided by 10 at epochs 150 and 250.
We scale the learning rate by the active number of workers based on the linear scaling rule found in~\cite{goyal2017accurate}.
The momentum parameter~($\beta$) is set to 0.9, and weight decay is set to $10^{-4}$.

We vary the number of workers from one to eight.
Each worker has one NVIDIA Tesla K80 with 12GB of memory, a quad-core Intel Xeon processor running at 2.2GHz with 16GB of DRAM, and a 10Gbit NIC.


\subsection{Results}

\begin{table*} 
\centering
  \caption{Test accuracy of models trained using our various SGD configurations with various number of workers~($W$).}
  \label{tab:accuracy}
   \resizebox{0.9\textwidth}{!}{%
   \def\arraystretch{1.15}
   \begin{tabular}{c|c c c c|c c c c c}
    
    \multirow{2}{*}{} & \multicolumn{4}{c|}{Layer-wise sparsification scope} & \multicolumn{4}{c}{Global sparsification scope} \\ 
  & \textbf{$W=1$} & \textbf{$W=2$} & \textbf{$W=4$} & \textbf{$W=8$} & \textbf{$W=1$} & \textbf{$W=2$} & \textbf{$W=4$} & \textbf{$W=8$}  \\
  \hline \hline

  Standard SGD & 94.50\% & 94.53\% & 94.40\% & 94.30\% & 94.50\% & 94.53\% & 94.40\% & 94.30\% \\
  Top-k & 93.63\% & 93.70\% & 93.68\% & 93.41\% & 91.45\% & 91.51\% & 91.75\% & 91.50\% \\
  Random-k ($allGather$) & 93.43 \% & 93.30\% & 93.12\% & 92.54\% & 91.73\% & 91.76\% & 91.16\% & 91.45\% \\
  Random-k ($allReduce$) & 93.43\% & 93.20\% & 93.15\% & 92.43\% & 91.73\% & 91.66\% & 91.26\% & 91.25\% \\
  Block-random-k ($allGather$) & 93.16\% & 93.10\% & 93.12\% & 93.10\% & 91.18\% & 91.05\% & 91.11\% & 90.45\% \\
  Block-random-k ($allReduce$) & 93.16\% & 92.50\% & 90.80\% & 90.63\% & 91.18\% & 88.20\% & 79.00\% & 78.11\% \\
  \hline \hline
  \end{tabular}
  }
\end{table*}

\subsubsection{Test Accuracy}
Table~\ref{tab:accuracy} depicts the test accuracy of our various configurations.
The first thing we notice is that configurations with sparsification scope of a single layer perform better compared to the configurations with the global sparsification scope, regardless of the sparsification or communication schemes and number of workers.
We believe the reason behind this trend is that in the layer-wise case, all layers participate in moving toward the optimal solution at each iteration, as gradients from all layers are shared among the workers.
For the configurations with global sparsification scope however, there is no guarantee that gradients from all layers will be shared.
As each layer may carry different information about the model, using gradients of all layers means we use more diverse information to update parameters, which leads to better performance.


        

\begin{table*}

    \caption{Breakdown of time spent in one training step in the layer-wise case while using 8 GPUs. \\
    \protect \tikz{\fill[fill=harvardcrimson] (0.0,0) rectangle (0.2,0.2)} Forward pass,
    \protect \tikz{\fill[fill=hanblue] (0.0,0) rectangle (0.2,0.2)} Backward pass, 
    \protect \tikz{\fill[fill=forestgreen] (0.0,0) rectangle (0.2,0.2)} Gradient exchange, 
    \protect \tikz{\fill[pattern=north east lines, pattern color=black!70!forestgreen] (0.0,0) rectangle (0.2,0.2)} Coding and Decoding.}
    
\begin{tabularx}{0.9\textwidth}{Xl c}

\\
    &  Time breakdown per batch & Time (ms)\\
    \hline
    Standard SGD 
        & \tikz{
        \fill[fill=harvardcrimson] (0.0,0) rectangle (0.19989235378031386,0.2);
        \fill[fill=hanblue] (0.19989235378031386,0) rectangle (0.633104536376605,0.2);
        \fill[fill=forestgreen] (0.633104536376605,0) rectangle (2.679500920114123,0.2);
        \fill[pattern=north east lines, pattern color=black!70!forestgreen] (0.633104536376605,0) rectangle (0.6567747503566337,0.2);
    } & 375\\
    Top-k 
        & \tikz{
        \fill[fill=harvardcrimson] (0.0,0) rectangle (0.19525523537803138,0.2);
        \fill[fill=hanblue] (0.19525523537803138,0) rectangle (0.613875092724679,0.2);
        \fill[fill=forestgreen] (0.613875092724679,0) rectangle (4.139841233951498,0.2);
        \fill[pattern=north east lines, pattern color=black!70!forestgreen] (0.613875092724679,0) rectangle (3.929031875891584,0.2);
        
    } & 580\\ 
    Block-random-k ($AllGather$) 
        & \tikz{
        \fill[fill=harvardcrimson] (0.0,0) rectangle (0.19586848074179744,0.2);
        \fill[fill=hanblue] (0.19586848074179744,0) rectangle (0.6125451711840229,0.2);
        \fill[fill=forestgreen] (0.6125451711840229,0) rectangle (2.224333659058488,0.2);
        \fill[pattern=north east lines, pattern color=black!70!forestgreen] (0.6125451711840229,0) rectangle (2.1201644864479317,0.2);
        
    } & 311\\
    Block-random-k ($AllReduce$)
        & \tikz{
        \fill[fill=harvardcrimson] (0.0,0) rectangle (0.19382674750356635,0.2);
        \fill[fill=hanblue] (0.19382674750356635,0) rectangle (0.614403209700428,0.2);
        \fill[fill=forestgreen] (0.614403209700428,0) rectangle (1.9499590228245367,0.2);
        \fill[pattern=north east lines, pattern color=black!70!forestgreen] (0.614403209700428,0) rectangle (1.8709612981455068,0.2);
        
    } & 273\\
    Randomk ($AllGather$) 
        & \tikz{
        \fill[fill=harvardcrimson] (0.0,0) rectangle (0.19664681169757492,0.2);
        \fill[fill=hanblue] (0.19664681169757492,0) rectangle (0.6186662981455064,0.2);
        \fill[fill=forestgreen] (0.6186662981455064,0) rectangle (4.392142439372326,0.2);
        \fill[pattern=north east lines, pattern color=black!70!forestgreen] (0.6186662981455064,0) rectangle (4.165998723252497,0.2);
        
    } & 615\\
    Randomk ($AllReduce$) 
        & \tikz{
        \fill[fill=harvardcrimson] (0.0,0) rectangle (0.19703982881597717,0.2);
        \fill[fill=hanblue] (0.19703982881597717,0) rectangle (0.6200526961483596,0.2);
        \fill[fill=forestgreen] (0.6200526961483596,0) rectangle (3.0665606276747504,0.2);
        \fill[pattern=north east lines, pattern color=black!70!forestgreen] (0.6200526961483596,0) rectangle (2.9567850213980025,0.2);
        
    } & 429\\
        & \tikz{
        \draw[gray] (0,0) -- (5,0);
        \draw[gray] (0.0,-2pt) -- (0.0,2pt);
        \draw[gray] (0.7132667617689016,-2pt) -- (0.7132667617689016,2pt)node[anchor=north] {\tiny$0.1$};
        \draw[gray] (1.4265335235378032,-2pt) -- (1.4265335235378032,2pt)node[anchor=north] {\tiny$0.2$};
        \draw[gray] (2.139800285306705,-2pt) -- (2.139800285306705,2pt)node[anchor=north] {\tiny$0.3$};
        \draw[gray] (2.8530670470756063,-2pt) -- (2.8530670470756063,2pt)node[anchor=north] {\tiny$0.4$};
        \draw[gray] (3.566333808844508,-2pt) -- (3.566333808844508,2pt)node[anchor=north] {\tiny$0.5$};
        \draw[gray] (4.27960057061341,-2pt) -- (4.27960057061341,2pt)node[anchor=north] {\tiny$0.6$};
        \draw[gray] (4.992867332382312,-2pt) -- (4.992867332382312,2pt)node[anchor=north] {\tiny$0.7$};
    } \\
    \hline

\end{tabularx}

    \label{tab:time_breakdown}

\end{table*}

Second, we notice the top-k configuration outperforms the other two sparsification schemes across the board, because it chooses more valuable gradients to share across the workers.
Random-k and block-random-k~(\verb|allGather|) have comparable performance when layer-wise sparsification is used, while the random-k performs better than block-random-k when we use the global scope for sparsification.
We believe this is because using block-random-k with the global scope configuration leads to only a few number of layers to be involved in updating model parameters.

The results also show there is no considerable difference between the test accuracy of the random-k algorithm using \verb|allGather| and \verb|allReduce| aggregation schemes.
However, for the block-random-k algorithm, the test accuracy drops significantly with more workers, especially when \verb|allReduce| is used.
We believe the reason behind this effect is similar to why layer-wise sparsification performs better than global spasification, which is using more diverse information.
The accuracy decreases for all configurations with more workers.
This is because the global batch size increases with the number of workers, which leads to the generalization gap problem~\cite{lin2018don}.

\subsubsection{Execution Time}

Table~\ref{tab:time_breakdown} shows a detailed breakdown of the time spent on each training step (mini-batch) into forward pass, backward pass, gradient exchange, and compression/decompression.
The time spent in the forward and backward passes is constant across all algorithms.
As standard SGD does not have any gradient compression, the time spent on exchanging gradients is the communication time among workers.
However, the rest of the configurations need some time to compress/decompress the gradients before/after the exchange.

The results show that even though gradient compression leads to less time being spend in the exchanging gradients, the compression/decompression overhead is too much to get any overall benefits.
Block-random-k is the only configuration that has lower execution time than standard SGD.
This is because of the simplicity of block-random-k compared to the other two sparsification schemes; the top-k configuration spends most of its time to find the top k gradients, and random-k has the extra overhead for random memory accesses.
We believe the benefits of gradient compression will be much bigger with more workers, as standard SGD has to communicate all the gradients across all the workers, but we could not show this effect due to limited available resources.



\section{Conclusion}

Gradient compression is a promising method for addressing the communication bottleneck in distributed synchronous SGD.
However, this approach has not been widely adopted in practice as the existing compression schemes either run slower than SGD or do not reach the same test performance.
In this paper, we investigate several sparsification schemes and discuss different aspects of sparsified SGD.
We show that layer-wise sparsification performs better than the global case.
We also introduce block-random-k, a new efficient sparsification scheme, which is simple yet practical.
This scheme is faster than standard SGD and has a comparable test accuracy to SGD.

\section*{Acknowledgments}
We thank Lie He, Sebastian U. Stich, and Arash Pourhabibi for their valuable feedback and suggestions.
This work was supported by Google Cloud credits.




\bibliography{references} 

\begin{thebibliography}{29}
\providecommand{\natexlab}[1]{#1}
\providecommand{\url}[1]{\texttt{#1}}
\expandafter\ifx\csname urlstyle\endcsname\relax
  \providecommand{\doi}[1]{doi: #1}\else
  \providecommand{\doi}{doi: \begingroup \urlstyle{rm}\Url}\fi

\bibitem[Alistarh et~al.(2016)Alistarh, Li, Tomioka, and
  Vojnovic]{alistarh2016qsgd}
Alistarh, D., Li, J., Tomioka, R., and Vojnovic, M.
\newblock Qsgd: Randomized quantization for communication-optimal stochastic
  gradient descent.
\newblock \emph{arXiv preprint arXiv:1610.02132}, 2016.

\bibitem[Alistarh et~al.(2018)Alistarh, Hoefler, Johansson, Konstantinov,
  Khirirat, and Renggli]{alistarh2018convergence}
Alistarh, D., Hoefler, T., Johansson, M., Konstantinov, N., Khirirat, S., and
  Renggli, C.
\newblock The convergence of sparsified gradient methods.
\newblock In \emph{Advances in Neural Information Processing Systems}, pp.\
  5973--5983, 2018.

\bibitem[Chiu et~al.(2018)Chiu, Sainath, Wu, Prabhavalkar, Nguyen, Chen,
  Kannan, Weiss, Rao, Gonina, et~al.]{chiu2018state}
Chiu, C.-C., Sainath, T.~N., Wu, Y., Prabhavalkar, R., Nguyen, P., Chen, Z.,
  Kannan, A., Weiss, R.~J., Rao, K., Gonina, E., et~al.
\newblock State-of-the-art speech recognition with sequence-to-sequence models.
\newblock In \emph{2018 IEEE International Conference on Acoustics, Speech and
  Signal Processing (ICASSP)}, pp.\  4774--4778. IEEE, 2018.

\bibitem[Das et~al.(2016)Das, Avancha, Mudigere, Vaidynathan, Sridharan,
  Kalamkar, Kaul, and Dubey]{das2016distributed}
Das, D., Avancha, S., Mudigere, D., Vaidynathan, K., Sridharan, S., Kalamkar,
  D., Kaul, B., and Dubey, P.
\newblock Distributed deep learning using synchronous stochastic gradient
  descent.
\newblock \emph{arXiv preprint arXiv:1602.06709}, 2016.

\bibitem[Dean et~al.(2012)Dean, Corrado, Monga, Chen, Devin, Mao, Ranzato,
  Senior, Tucker, Yang, et~al.]{dean2012large}
Dean, J., Corrado, G., Monga, R., Chen, K., Devin, M., Mao, M., Ranzato, M.,
  Senior, A., Tucker, P., Yang, K., et~al.
\newblock Large scale distributed deep networks.
\newblock In \emph{Advances in neural information processing systems}, pp.\
  1223--1231, 2012.

\bibitem[Devlin et~al.(2018)Devlin, Chang, Lee, and Toutanova]{devlin2018bert}
Devlin, J., Chang, M.-W., Lee, K., and Toutanova, K.
\newblock Bert: Pre-training of deep bidirectional transformers for language
  understanding.
\newblock \emph{arXiv preprint arXiv:1810.04805}, 2018.

\bibitem[Goyal et~al.(2017)Goyal, Doll{\'a}r, Girshick, Noordhuis, Wesolowski,
  Kyrola, Tulloch, Jia, and He]{goyal2017accurate}
Goyal, P., Doll{\'a}r, P., Girshick, R., Noordhuis, P., Wesolowski, L., Kyrola,
  A., Tulloch, A., Jia, Y., and He, K.
\newblock Accurate, large minibatch sgd: training imagenet in 1 hour.
\newblock \emph{arXiv preprint arXiv:1706.02677}, 2017.

\bibitem[He et~al.(2016{\natexlab{a}})He, Zhang, Ren, and Sun]{He_2016_CVPR}
He, K., Zhang, X., Ren, S., and Sun, J.
\newblock Deep residual learning for image recognition.
\newblock In \emph{The IEEE Conference on Computer Vision and Pattern
  Recognition (CVPR)}, June 2016{\natexlab{a}}.

\bibitem[He et~al.(2016{\natexlab{b}})He, Zhang, Ren, and Sun]{he2016deep}
He, K., Zhang, X., Ren, S., and Sun, J.
\newblock Deep residual learning for image recognition.
\newblock In \emph{Proceedings of the IEEE conference on computer vision and
  pattern recognition}, pp.\  770--778, 2016{\natexlab{b}}.

\bibitem[Iandola et~al.(2016)Iandola, Moskewicz, Ashraf, and
  Keutzer]{iandola2016firecaffe}
Iandola, F.~N., Moskewicz, M.~W., Ashraf, K., and Keutzer, K.
\newblock Firecaffe: near-linear acceleration of deep neural network training
  on compute clusters.
\newblock In \emph{Proceedings of the IEEE Conference on Computer Vision and
  Pattern Recognition}, pp.\  2592--2600, 2016.

\bibitem[Karimireddy et~al.(2019)Karimireddy, Rebjock, Stich, and
  Jaggi]{karimireddy2019error}
Karimireddy, S.~P., Rebjock, Q., Stich, S.~U., and Jaggi, M.
\newblock Error feedback fixes signsgd and other gradient compression schemes.
\newblock \emph{arXiv preprint arXiv:1901.09847}, 2019.

\bibitem[Koloskova et~al.(2019)Koloskova, Stich, and
  Jaggi]{koloskova2019decentralized}
Koloskova, A., Stich, S.~U., and Jaggi, M.
\newblock Decentralized stochastic optimization and gossip algorithms with
  compressed communication.
\newblock \emph{arXiv preprint arXiv:1902.00340}, 2019.

\bibitem[Krizhevsky et~al.(2009)Krizhevsky, Hinton,
  et~al.]{krizhevsky2009learning}
Krizhevsky, A., Hinton, G., et~al.
\newblock Learning multiple layers of features from tiny images.
\newblock 2009.

\bibitem[Li et~al.(2020)Li, Sahu, Talwalkar, and Smith]{li2020federated}
Li, T., Sahu, A.~K., Talwalkar, A., and Smith, V.
\newblock Federated learning: Challenges, methods, and future directions.
\newblock \emph{IEEE Signal Processing Magazine}, 37\penalty0 (3):\penalty0
  50--60, 2020.

\bibitem[Lin et~al.(2018)Lin, Stich, Patel, and Jaggi]{lin2018don}
Lin, T., Stich, S.~U., Patel, K.~K., and Jaggi, M.
\newblock Don't use large mini-batches, use local sgd.
\newblock \emph{arXiv preprint arXiv:1808.07217}, 2018.

\bibitem[Lin et~al.(2017)Lin, Han, Mao, Wang, and Dally]{lin2017deep}
Lin, Y., Han, S., Mao, H., Wang, Y., and Dally, W.~J.
\newblock Deep gradient compression: Reducing the communication bandwidth for
  distributed training.
\newblock \emph{arXiv preprint arXiv:1712.01887}, 2017.

\bibitem[McMahan et~al.(2016)McMahan, Moore, Ramage, Hampson,
  et~al.]{mcmahan2016communication}
McMahan, H.~B., Moore, E., Ramage, D., Hampson, S., et~al.
\newblock Communication-efficient learning of deep networks from decentralized
  data.
\newblock \emph{arXiv preprint arXiv:1602.05629}, 2016.

\bibitem[MLBench()]{MLBench_benchmark}
MLBench.
\newblock {Distributed Machine Learning Benchmark}.
\newblock URL \url{https://mlbench.github.io}.

\bibitem[Paszke et~al.(2019)Paszke, Gross, Massa, Lerer, Bradbury, Chanan,
  Killeen, Lin, Gimelshein, Antiga, Desmaison, Kopf, Yang, DeVito, Raison,
  Tejani, Chilamkurthy, Steiner, Fang, Bai, and Chintala]{NEURIPS2019_9015}
Paszke, A., Gross, S., Massa, F., Lerer, A., Bradbury, J., Chanan, G., Killeen,
  T., Lin, Z., Gimelshein, N., Antiga, L., Desmaison, A., Kopf, A., Yang, E.,
  DeVito, Z., Raison, M., Tejani, A., Chilamkurthy, S., Steiner, B., Fang, L.,
  Bai, J., and Chintala, S.
\newblock Pytorch: An imperative style, high-performance deep learning library.
\newblock In Wallach, H., Larochelle, H., Beygelzimer, A., d'~Alch\'{e}-Buc,
  F., Fox, E., and Garnett, R. (eds.), \emph{Advances in Neural Information
  Processing Systems 32}, pp.\  8024--8035. Curran Associates, Inc., 2019.

\bibitem[Renggli et~al.(2018)Renggli, Alistarh, Hoefler, and
  Aghagolzadeh]{renggli2018sparcml}
Renggli, C., Alistarh, D., Hoefler, T., and Aghagolzadeh, M.
\newblock Sparcml: High-performance sparse communication for machine learning.
\newblock \emph{arXiv preprint arXiv:1802.08021}, 2018.

\bibitem[Seide et~al.(2014)Seide, Fu, Droppo, Li, and Yu]{seide20141}
Seide, F., Fu, H., Droppo, J., Li, G., and Yu, D.
\newblock 1-bit stochastic gradient descent and its application to
  data-parallel distributed training of speech dnns.
\newblock In \emph{Fifteenth Annual Conference of the International Speech
  Communication Association}, 2014.

\bibitem[Stich et~al.(2018)Stich, Cordonnier, and Jaggi]{stich2018sparsified}
Stich, S.~U., Cordonnier, J.-B., and Jaggi, M.
\newblock Sparsified sgd with memory.
\newblock In \emph{Advances in Neural Information Processing Systems}, pp.\
  4452--4463, 2018.

\bibitem[Strom(2015)]{strom2015scalable}
Strom, N.
\newblock Scalable distributed dnn training using commodity gpu cloud
  computing.
\newblock In \emph{Sixteenth Annual Conference of the International Speech
  Communication Association}, 2015.

\bibitem[Vogels et~al.(2019)Vogels, Karimireddy, and Jaggi]{vogels2019powersgd}
Vogels, T., Karimireddy, S.~P., and Jaggi, M.
\newblock Powersgd: Practical low-rank gradient compression for distributed
  optimization.
\newblock \emph{arXiv preprint arXiv:1905.13727}, 2019.

\bibitem[Wangni et~al.(2018)Wangni, Wang, Liu, and Zhang]{wangni2018gradient}
Wangni, J., Wang, J., Liu, J., and Zhang, T.
\newblock Gradient sparsification for communication-efficient distributed
  optimization.
\newblock In \emph{Advances in Neural Information Processing Systems}, pp.\
  1299--1309, 2018.

\bibitem[Wen et~al.(2017)Wen, Xu, Yan, Wu, Wang, Chen, and Li]{wen2017terngrad}
Wen, W., Xu, C., Yan, F., Wu, C., Wang, Y., Chen, Y., and Li, H.
\newblock Terngrad: Ternary gradients to reduce communication in distributed
  deep learning.
\newblock In \emph{Advances in neural information processing systems}, pp.\
  1509--1519, 2017.

\bibitem[Zhang et~al.(2019{\natexlab{a}})Zhang, Chen, Mokhtari, Hassani, and
  Karbasi]{zhang2019quantized}
Zhang, M., Chen, L., Mokhtari, A., Hassani, H., and Karbasi, A.
\newblock Quantized frank-wolfe: Communication-efficient distributed
  optimization.
\newblock \emph{arXiv preprint arXiv:1902.06332}, 2019{\natexlab{a}}.

\bibitem[Zhang et~al.(2019{\natexlab{b}})Zhang, Yao, Sun, and
  Tay]{zhang2019deep}
Zhang, S., Yao, L., Sun, A., and Tay, Y.
\newblock Deep learning based recommender system: A survey and new
  perspectives.
\newblock \emph{ACM Computing Surveys (CSUR)}, 52\penalty0 (1):\penalty0 1--38,
  2019{\natexlab{b}}.

\bibitem[Zhou et~al.(2016)Zhou, Wu, Ni, Zhou, Wen, and Zou]{zhou2016dorefa}
Zhou, S., Wu, Y., Ni, Z., Zhou, X., Wen, H., and Zou, Y.
\newblock Dorefa-net: Training low bitwidth convolutional neural networks with
  low bitwidth gradients.
\newblock \emph{arXiv preprint arXiv:1606.06160}, 2016.

\end{thebibliography}
\bibliographystyle{icml2020}

\end{document}